\documentclass{article}
\usepackage{spconf,amsmath,graphicx}
\usepackage{multirow}
\usepackage{subfigure}
\usepackage{graphicx}
\usepackage{cite}
\usepackage{algpseudocode}
\usepackage{amsmath}
\usepackage{makecell}

\title{Improving CTC-based ASR Models with Gated Interlayer Collaboration}
\name{Yuting Yang, Yuke Li$^*$\thanks{$^*$ Corresponding author}, Binbin Du}
\address{NetEase Yidun AI Lab, Hangzhou, China \\
{\small {$\left\{yangyuting04, liyuke, dubinbin\right\}$@corp.netease.com} }
}

\begin{document}
%
\maketitle
\begin{abstract}
The CTC-based automatic speech recognition (ASR) models without the external language model usually lack the capacity to model conditional dependencies and textual interactions. In this paper, we present a \textbf{G}ated \textbf{I}nterlayer \textbf{C}ollaboration (\textbf{GIC}) mechanism to improve the performance of CTC-based models, which introduces textual information into the model and thus relaxes the conditional independence assumption of CTC-based models. Specifically, we consider the \textit{weighted} sum of token embeddings as the textual representation for each position, where the position-specific weights are the softmax probability distribution constructed via inter-layer auxiliary CTC losses. The textual representations are then fused with acoustic features by developing a gate unit. Experiments on AISHELL-1 \cite{bu2017aishell}, TEDLIUM2 \cite{rousseau2014enhancing}, and AIDATATANG \cite{aidatantang} corpora show that the proposed method outperforms several strong baselines.
\end{abstract}
\section{Introduction}
\label{sec:intro}
The end-to-end automatic speech recognition (E2E ASR) system has become increasingly popular due to its simpler architecture and outstanding performance. A variety of E2E ASR models have been explored in the literature, which can be categorized into three main approaches: attention-based encoder-decoder systems \cite{chan2015listen,bahdanau2016end,vaswani2017attention,dong2018speech}, transducer models \cite{graves2012sequence,battenberg2017exploring,zhang2020transformer}, and Connectionist Temporal Classification (CTC) models \cite{2006Connectionist,hannun2014deep,graves2014towards,miao2015eesen,kriman2020quartznet,majumdar2021citrinet}. The attention-based systems solve the ASR problem as a sequence mapping from speech feature sequences to texts using an encoder-decoder architecture. The transducer network extends CTC by additionally modeling the dependencies between outputs at different timesteps. The CTC method introduces the blank label to identify the no-output positions and utilizes frame-wise label sequences to obtain the alignments between the input and output sequences. In contrast to the attention-based and the transducer-based models, the CTC models only involve with the encoder module and naturally provide a non-autoregressive manner to generate sentences in a fast and straightforward way.

However, both the attention-based and transducer-based models have often shown better performance than the CTC-based model equipped without the external language model (LM) \cite{battenberg2017exploring}. The inferior performance of a CTC system stems from two aspects. First, CTC imposes the conditional independence constraint that the output prediction at each position is independent of each other, which is illogical for ASR tasks. Second, CTC only relies on the acoustic feature to generate sentences, while attention-based models utilize both the acoustic and textual features.

In this study, we focus on improving the performance of CTC-based models by addressing the issues mentioned above. To achieve this, we need to consider two problems: (1) \textit{how to introduce the textual information into the models during the training process}, and (2) \textit{how to fuse both the textual and the acoustic features}. We present a \textbf{G}ated \textbf{I}nterlayer \textbf{C}ollaboration (\textbf{GIC}) mechanism, which consists of (1) an $embedding$ layer to introduce the textual information into the CTC-based model by utilizing the sequence of soft labels of intermediate layer predictions, and (2) a gate unit to fuse the textual and acoustic features. Specifically, we train the model with the intermediate auxiliary CTC loss calculated by the interlayer output of the model. The probability distributions of the intermediate layers naturally serve as soft labels, which are also used to obtain the textual embedding at each position. We conduct extensive experiments to compare our methods with previous works on various ASR datasets, including AISHELL-1, AIDATATANG, and TEDLIUM2. GIC sets the new state-of-the-art among several strong baselines. Specifically, our method achieves 4.3\% on the AISHELL-1 test set, 4.1\% on the AIDATATANG test set, and 7.1\% on the TEDLIUM2 test set, respectively. These improvements demonstrate the effectiveness of introducing the GIC mechanism into CTC-based models. 

\section{Related work}
\label{sec:related_work}
\begin{figure}[t]
  \centering
  \includegraphics[height=7cm,width=7.5cm]{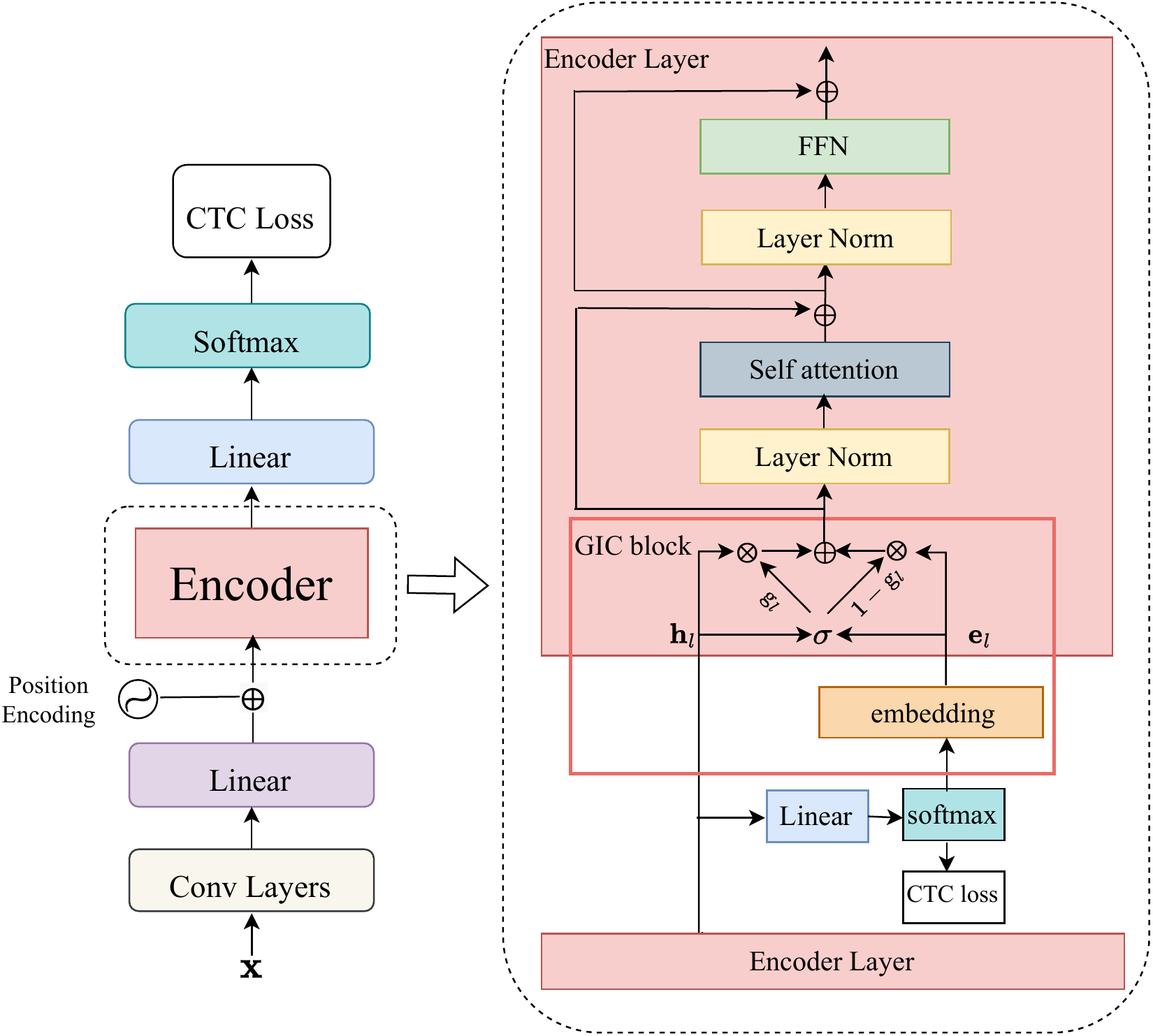}
  \caption{Illustration of our method. We compute intermediate CTC losses for some particular encoder layers. Then, GIC blocks introduce the textual information into the model generating by intermediate predictions.}
  \label{fig:fig1}
   \vspace{-0.3cm} 
\end{figure}

Many works have explored the improvements of auxiliary tasks for CTC models \cite{2018Hierarchical,krishna2018hierarchical,rao2017multi,rao2017exploring,lee2021intermediate,nozaki2021relaxing,higuchi2022hierarchical,fujita2022multi}. Intermediate CTC \cite{lee2021intermediate} utilize intermediate auxiliary CTC losses of the encoder module to improve the performance of CTC-based models. Self-conditioning CTC \cite{nozaki2021relaxing} and hierarchical conditional CTC \cite{higuchi2022hierarchical} improve the performance of CTC-based models by conditioning the final prediction on the intermediate predictions. In addition, integrating the external LM for shallow fusion brings performance improvements for CTC-based models \cite{hannun2014deep,graves2014towards}. 
Our method differs from the previous literature \cite{rao2017exploring,2018Hierarchical,krishna2018hierarchical,rao2017multi,lee2021intermediate,nozaki2021relaxing,higuchi2022hierarchical,fujita2022multi} in a number of ways, including the method of introducing textual features into the CTC model and the method of gate-based feature fusion. 

\section{Proposed Method}
\label{sec:proposed}

In the training process, we use the feature sequence $X=\left\{x_1,x_2,...,x_T\right\}$ and the text sequence $Y=\left\{y_1,y_2,...,y_n\right\}$ to train the network parameters. Figure~\ref{fig:fig1} shows the framework of the proposed method, which takes Transformer as the backbone for example. In experiments, we adopt both the Transformer \cite{vaswani2017attention} and Conformer \cite{gulati2020conformer} as backbones for comparison with strong baselines.  

\subsection{Baseline Model Architecture}

\textbf{Transformer encoder.} An encoder layer of the Transformer \cite{vaswani2017attention} comprises two main sub-layers: a self-attention layer followed by a feed-forward network. Formally,
\begin{equation}
\begin{split}
{\rm h}_{l}^{\prime} &= {\rm h}_{l}^{in} + {\rm SAN}({\rm Q}_l,{\rm K}_l,{\rm V}_l), \\
{\rm h}_{l+1}^{in} &= {\rm h}_{l} =  {\rm h}_l^{\prime} +  {\rm FFN}({\rm LN}({\rm h}_l^{\prime})),
\end{split}
\label{eq1}
\end{equation}
where ${\rm h}_l^{in}$ and ${\rm h}_l$ are the input and output of $l$-th layer. $\rm LN(\cdot)$, $\rm SAN(\cdot)$ and $\rm FFN(\cdot)$ are layer normalization \cite{ba2016layer}, attention mechanism, and feed-forward networks respectively. (${\rm Q}_l$, ${\rm K}_l$, ${\rm V}_l$) are query, key and value vectors that are transformed from the normalized $l$-th encoder layer input, that is ${\rm LN}({\rm h}_{l}^{in})$.

\textbf{Conformer encoder.} Conformer \cite{gulati2020conformer} combines Transformer and convolutional neural network layers to efficiently learn both global and local representations. An encoder block of Conformer is composed of four modules, i.e., a feed-forward module, a self-attention module, a convolution module, and a second feed-forward module. Formally,
\begin{equation}
\begin{split}
{\rm h}_l^{\prime} &= {\rm h}_{l}^{in} +\frac{1}{2} {\rm FFN}({\rm h}_{l}^{in}),\\
{\rm s}_l &= {\rm h}_{l}^{\prime} + {\rm SAN}({\rm Q}_l,{\rm K}_l,{\rm V}_l),\\
{\rm o}_{l} &= {\rm s}_{l} + {\rm Convolution}({\rm s}_{l}),\\
{\rm h}_{l+1}^{in} &= {\rm h}_{l} = {\rm LN}({\rm o}_{l} +\frac{1}{2}{\rm FFN}({\rm o}_{l})).
\end{split}
\label{eq2}
\end{equation}

\textbf{Connectionist Temporal Classification.} CTC \cite{2006Connectionist} optimizes the model by maximizing the likelihood of all valid frame-wise alignments $\beta^{-1}(Y)$. Formally,
\begin{equation}
\begin{split}
  {\rm q}_L &= {\rm Softmax}({\rm Linear}_{d_{m}\rightarrow{|V|}}({\rm LN}({\rm h}_{L}))),  \\
  P_{ctc}(Y|{\rm h}_{L}) &= \sum_{\alpha \in{\beta^{-1}(Y)}}\prod_{t=1}^{T}{\rm q}_L^{t}[\alpha_t],
\label{eq3}
\end{split}
\end{equation}
where ${\rm q}_L$ is the probability distribution of each frame over the vocabulary $V$, and $d_m$ is the dimension of the attention layer. ${\rm q}_L^{t}$ denotes the $t$-th column of ${\rm q}_L$, and ${\rm q}_L^{t}[\alpha_t]$ denotes the $\alpha_t$-th symbol of ${\rm q}_L^{t}$. ${\rm h}_L$ is the output of the encoder module, and $L$ indicates the number of encoder layers. The CTC loss is defined as the negative log-likelihood as:
\begin{equation}
  \pounds_{ctc} = - \log\ P_{ctc}(Y|{\rm h}_{L}).
  \label{eq4}
\end{equation}

\textbf{Intermediate CTC.} Intermediate CTC \cite{lee2021intermediate} utilizes the representations of intermediate encoder layers to calculate the auxiliary CTC losses to further improve the performance of the models. We train the model with a total of $K$ intermediate layers to calculate the inter-layer CTC losses. 
The total loss function is a linear combination of the final layer CTC loss $\pounds_{ctc}$ (Eq.{\ref{eq4}}) and the intermediate CTC losses. Formally,
\begin{equation}
  \pounds = (1-{\lambda}) \pounds_{ctc} + {\lambda}\frac{1}{K}\sum_{i=1}^{K} \pounds_{\frac{iL}{K+1}}.
\label{eq18}
\end{equation}
The intermediate CTC loss $\pounds_l$ is calculated with the output of the $l$-th encoder layer (${\rm h}_l$) as  
\begin{equation}
\pounds_{l}= - \log\ P_{ctc}(Y|{\rm h}_{l}).
  \label{interctc}
\end{equation}
%

\subsection{Gated Interlayer Collaboration Mechanism}

\begin{figure}[htbp]
\centering
\subfigure[Effect of $K$]{
\begin{minipage}[t]{0.45\linewidth}
\centering
\includegraphics[height=2.5cm,width=4cm]{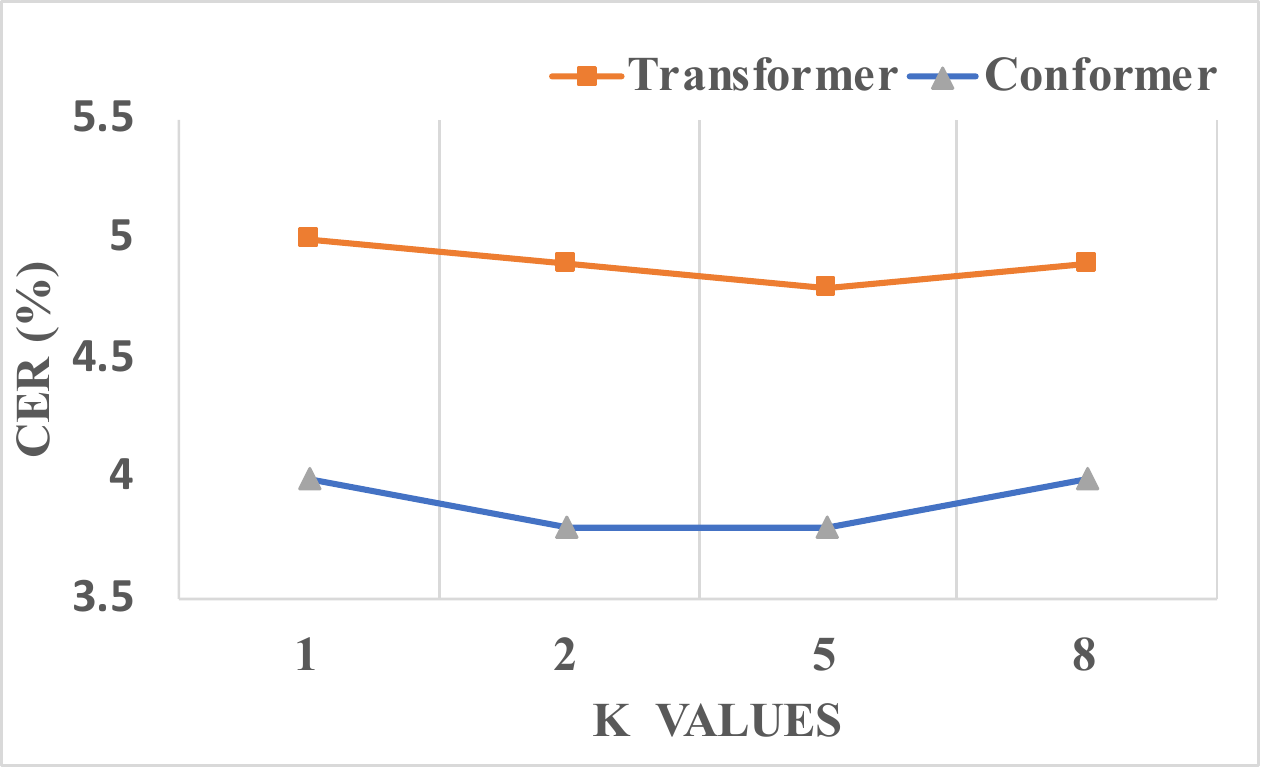}
\end{minipage}%
}%
\subfigure[Effect of $\lambda$]{
\begin{minipage}[t]{0.45\linewidth}
\centering
\includegraphics[height=2.5cm,width=4cm]{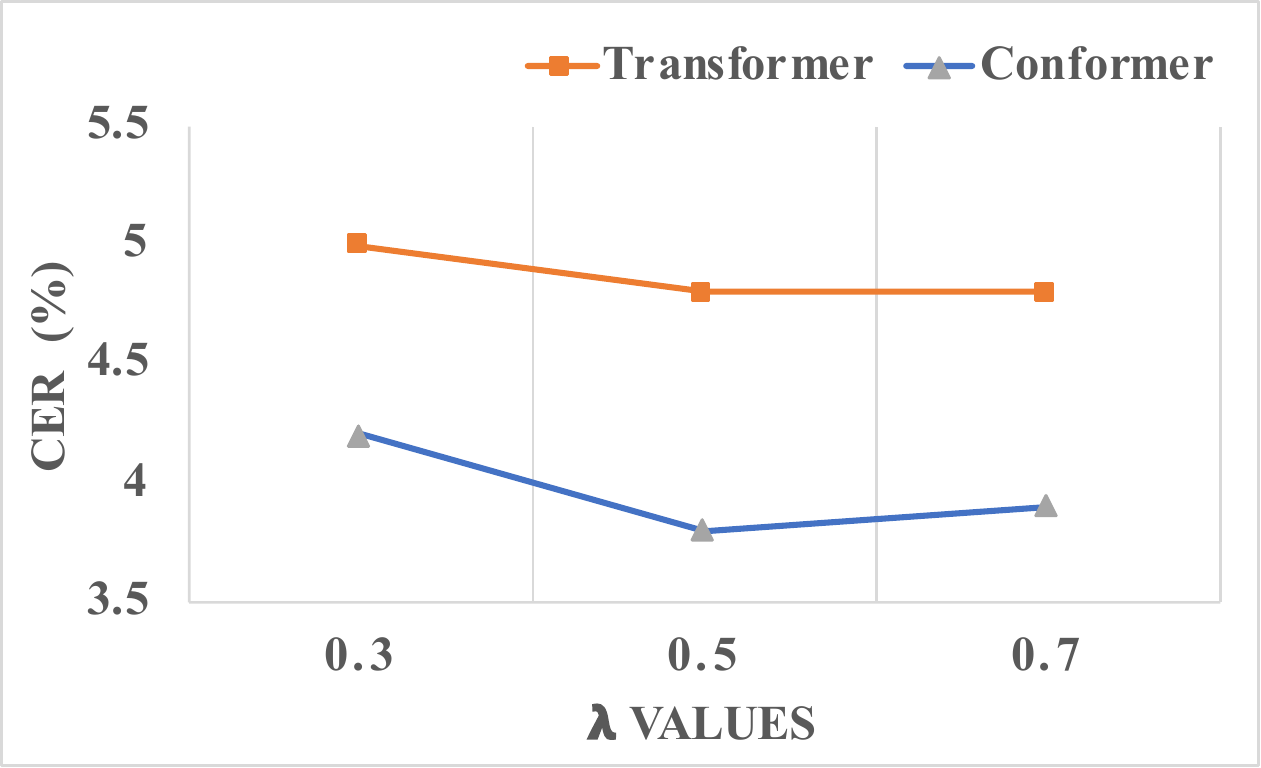}
\end{minipage}%
}%
\centering
\caption{Effect of $K$ and $\lambda$ on the AIDATATANG validation set. (a) demonstrates the performance of different $K$ values with $\lambda=0.5$. (b) demonstrates the performance of different $\lambda$ values under the optimal parameter of $K=5$.}
\label{fig:fig2}
\vspace{-0.2cm} 
\end{figure}

Figure~\ref{fig:fig1} shows the framework of the proposed method. We propose a gated interlayer collaboration (\textbf{GIC}) mechanism to make CTC-based models have access to textual information generating by intermediate predictions of the encoder module. The GIC block first introduces an $embedding$ layer that is used to get the position-specific textual representation by a weighted-combination of all token embeddings. Formally,
\begin{equation}
  {\rm e}_{l}^j = \sum_{i}^{|V|} {\rm q}_{l}^{(i,j)} \odot emb_i,
  \label{eq14}
\end{equation}
where $emb_i$ represents the $i$-th column of the embedding matrix. ${\rm q}_{l}^{(i,j)}$ indicates the $i$-th probability mass in the soft labels at the $j$-th frame, where the soft label is calculated by a softmax activation given the hidden state ${\rm h}_l^j$ of the $j$-th frame
\begin{equation}
  {\rm q}_{l} = {\rm Softmax}({\rm Linear}_{d_{m}\rightarrow{|V|}}({\rm LN}({\rm h}_{l}))).
  \label{eq13}
\end{equation}
In Eq.{\ref{eq13}}, $l$ is one of the particular intermediate layers and the output of the $l$-th layer ${\rm h}_l$ is used to calculate the intermediate CTC loss in Eq.{\ref{interctc}}.
Then, a gate unit combines the acoustic features ${\rm h}_{l}$ and the textual features ${\rm e}_{l}$ by a $sigmoid$ activation:
\begin{equation}
\begin{split}
  {\rm g}_{l} &= {\rm Sigmoid} (W_1 {\rm h}_{l} + W_2 {\rm e}_{l} + b), \\
    {\rm h}_{l+1}^{in} &= {\rm g}_{l} \odot {\rm h}_{l} + (1 - {\rm g}_{l})  \odot {\rm e}_{l}.
\end{split}
  \label{eq15}
\end{equation}
The gated combination ${\rm h}_{l+1}^{in}$ is then fed into the next encoder layer, and thus our approach benefits from both the textual and the acoustic interactions.

\section{Experiments}
\label{sec:experiments}

We conduct experiments on three ASR corpora: AISHELL-1 (178 hours, Chinese) \cite{bu2017aishell}, AIDATATANG (200 hours, Chinese) \cite{aidatantang}, and TEDLIUM2 (207 hours, English) \cite{rousseau2014enhancing}. We apply speed perturbation \cite{ko2015audio} and SpecAugment \cite{2019SpecAugment} to the training data. For all experiments, the input speech features are 80-dimensional filterbank (FBank) features with 3-dimensional pitch features computed on 25ms windows with 10ms shifts. The vocabulary includes 4231 characters for AISHELL-1, 3943 characters for AIDATATANG, and 500 subwords for TEDLIUM2, respectively.

\begin{table}[t]
\small
\renewcommand{\arraystretch}{1.1}
  \caption{Character error rate (CER) on AISHELL-1 with the \textbf{Conformer} backbone. $^\dag$ denotes the results from previous literature \cite{fujita2022multi}. }
  \label{tab:aishell-conformer}
  \centering
  \setlength\tabcolsep{1.2mm}{
  \begin{tabular}{lccccc}
  \Xhline{2\arrayrulewidth}
    \multirow{2}*{\bf
    Methods}&\multirow{2}*{\bf{\#Param (M)}}&\multicolumn{2}{c}{\bf  w/o LM }&\multicolumn{2}{c}{\bf w/ LM} \\
	&&\makecell[c]{\bf{dev}} &\makecell[c]{\bf{test}}&\makecell[c]{\bf{dev}} &\makecell[c]{\bf{test}} \\
	\hline
    Hybrid \cite{guo2021recent}&46&4.4&4.7&-&- \\
    \hline
    Transducer\cite{tian2021consistent}&89&4.3&4.6&4.1&4.4 \\
    \hline
    SC-CTC&-&4.3$^\dag$&4.6$^\dag$&-&- \\
    Alternate-CTC\cite{fujita2022multi}&-&4.2&4.5&-&-\\
    CTC &50&4.8&5.2&4.6&4.9 \\
    Intermediate-CTC &50&4.3&4.7&4.2&4.6\\
    SC-CTC &51&4.2&4.6&4.1&4.5\\
    \textbf{GIC (ours)}&52&\textbf{4.0}&\textbf{4.4}&\textbf{4.0}&\textbf{4.3}\\
  \Xhline{2\arrayrulewidth}
  \end{tabular}}
  \vspace{-0.2cm} 
\end{table}

\begin{table}[t]
\small
\renewcommand{\arraystretch}{1.1}
  \caption{ Word error rate (WER) on TEDLIUM2 with the \textbf{Conformer} backbone.
  $^\$/^\ddag$ denote the results from previous literature \cite{higuchi2021comparative}/\cite{higuchi2022hierarchical}. }
  \label{tab:ted-conformer}
  \centering
  \setlength\tabcolsep{3mm}{
  \begin{tabular}{lcc}
  \Xhline{2\arrayrulewidth}
    {\bf Methods}&\bf{Dev}&\bf{Test}\\
	\hline
	Hybrid &10.4$^\$$&8.4$^\$$\\
	\hline
	Transducer &8.6$^\$$&8.2$^\$$\\
	\hline
    Improved Mask-CTC \cite{higuchi2021improved} &-&8.6 \\
    \hline
    CTC&8.9$^\$$&8.6$^\$$\\
    Intermediate-CTC &8.5$^\$$&8.3$^\$$\\
    SC-CTC &8.5$^\ddag$&7.8$^\ddag$\\
    HC-CTC \cite{higuchi2022hierarchical}&{8.0}&7.6\\
    \textbf{GIC (ours)}&\textbf{7.9}&\textbf{7.3}\\
    \hspace{0.2cm}{\bf{+ beam search with a 4-gram LM}}&\textbf{7.7}&\textbf{7.1}\\
  \Xhline{2\arrayrulewidth}
  \end{tabular}}
  \vspace{-0.3cm}
\end{table}

\begin{figure}[t]
  \centering
  \includegraphics[height=3.2cm,width=6.0cm]{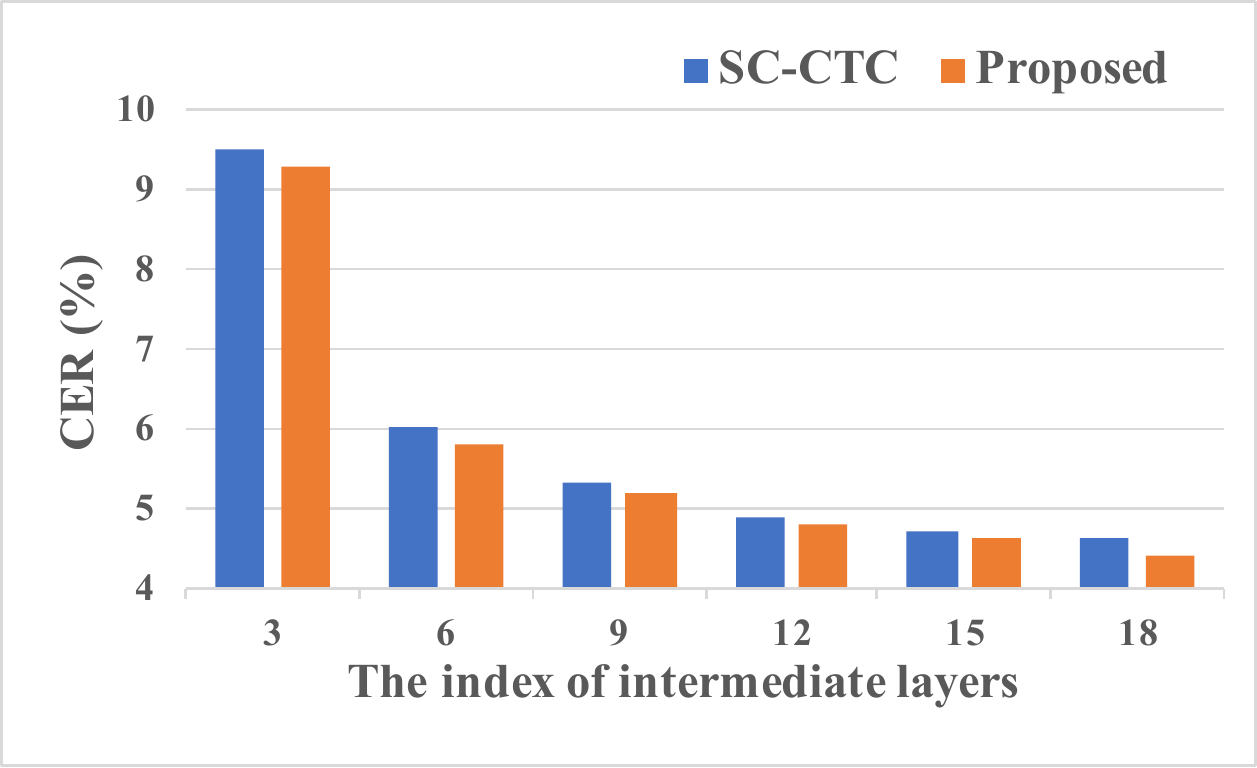}
  \caption{The CER comparison of intermediate layer outputs of the SC-CTC and the proposed models on the AISHELL-1 test set using Conformer networks with CTC greedy decoding. }
  \label{fig:fig3}
\end{figure}

\vspace{-0.05cm}
\subsection{Experimental Setup} 
We conducted experiments using ESPnet \cite{watanabe2018espnet}. We adopt two CNN-based downsampling layers followed by an 18-layer encoder for our experiments. For both Transformer and Conformer backbones, the dimension of the attention layer is 256 with 4 split heads, and the dimension of the feed-forward layer is 2048. Specifically, the kernel size is 15 for Conformer. We use a single-LSTM-layer decoder for the transducer-based models with the same 18-layer encoder. The Adam optimizer \cite{kingma2014adam} with 25000 warmup steps are used for training. We train the models for 100 epochs on both AISHELL-1 and TEDLIUM2, and 50 epochs on AIDATATANG. We conduct experiments to compare our method with previous CTC-based models under the same model scale, including CTC \cite{2006Connectionist}, intermediate-CTC training with the intermediate CTC loss \cite{lee2021intermediate}, and SC-CTC training with the intermediate CTC loss and the self-conditioning mechanism \cite{nozaki2021relaxing}. The results are obtained by either CTC greedy decoding without the external LM or beam search (beam size = 10) with a 4-gram LM. 

Figure~\ref{fig:fig2} shows the effect of different numbers of $K$ and different values of $\lambda$ in Eq.{\ref{eq18}}. We found that $K=5$ and $\lambda=0.5$ achieve better results. Therefore, we set $K=5$ (interlayer set of $\left\{3,6,9,12,15\right\}$) and $\lambda=0.5$ for our experiments.



\begin{table}[th]
\small
\renewcommand{\arraystretch}{1.1}
  \caption{Character error rate (CER) on AIDATATANG with the \textbf{Conformer} backbone.}
  \label{tab:aidatatang-conformer}
  \centering
\setlength\tabcolsep{4mm}{
  \begin{tabular}{lcccc}
  \hline
	    \multirow{2}*{\bf
    Methods}&\multicolumn{2}{c}{\bf  w/o LM }&\multicolumn{2}{c}{\bf w/ LM} \\
	&\makecell[c]{\bf{dev}} &\makecell[c]{\bf{test}}&\makecell[c]{\bf{dev}} &\makecell[c]{\bf{test}} \\
	\hline
    Hybrid\cite{guo2021recent}&4.3&5.0&-&-\\
    \hline
    Transducer&4.6&5.3&-&-\\
    \hline 
    CTC&4.9 &5.5&4.0&4.6\\
    \textbf{GIC (ours)}&\textbf{3.8}&\textbf{4.4}&\textbf{3.5}&\textbf{4.1}\\
  \hline
  \end{tabular}
  }
\vspace{-1em}
\end{table}

\vspace{-0.05cm} 
\subsection{Results} 
Table~\ref{tab:aishell-conformer} shows the results of different methods on AISHELL-1 based on the Conformer backbone. In addition to recently published CTC models \cite{fujita2022multi}, we compare our method with a few state-of-the-art models, including the hybrid CTC/Attention model \cite{guo2021recent} and the transducer-based method \cite{tian2021consistent}. Our model sets the start-of-the-art results among all Conformer-based models. Specifically, our method achieves CER of 4.0\%/4.4\% on AISHELL-1 dev/test sets using CTC greedy decoding without the external LM, outperforming the hybrid systems and other CTC models with slightly overhead parameters, and outperforming the transducer-based models while having the benefit of non-autoregressive decoding manner.
Furthermore, our method performs better to alleviate the conditional independence hypothesis of CTC-based models compared with the LM shallow fusion and the transducer network. Our method achieves CER of 4.0\%/4.3\% on AISHELL-1 dev/test sets with a 4-gram LM for shallow fusion. The external LM brings less boost since the proposed method introduced text interactions in the training process. 

Table~\ref{tab:ted-conformer} and Table~\ref{tab:aidatatang-conformer} show the results of different methods with the Conformer backbone on TEDLIUM2 and AIDATATANG, respectively. GIC achieves WER of 7.9\%/7.3\% with the CTC greedy search method and WER of 7.7\%/7.1\% with a 4-gram LM for shallow fusion on TEDLIUM2 dev/test sets, outperforming the results in recent literature \cite{higuchi2021comparative,higuchi2022hierarchical,higuchi2021improved}. The results show that GIC is also effective on the English dataset. Moreover, GIC achieves CER of 3.8\%/4.4\% on AIDATATANG dev/test sets without the external LM. 

Table~\ref{tab:transformer1} shows the results on AISHELL-1 and AIDATA-TANG datasets, using the Transformer backbone. Our method is also effective for the Transformer network and achieves the lowest CER (i.e., 5.1\% on the AISHELL-1 test set without LM) among all other Transformer-based models. 
\begin{table}[th]
\small
\renewcommand{\arraystretch}{1.1}
  \caption{Character error rate (CER) of \textbf{Transformer} backbone on AISHELL-1 and AIDATATANG without LM. $^\S/^\star$ denote the results from the previous literature \cite{nozaki2021relaxing}/\cite{guo2021recent}.}
  \label{tab:transformer1}
  \centering
  \setlength\tabcolsep{1.1mm}{
  \begin{tabular}{lccccc}
  \Xhline{2\arrayrulewidth}
	\multirow{2}*{\bf Methods}&\multirow{2}*{\bf{\#Param (M)}}&\multicolumn{2}{c}{\bf {AISHELL-1}}&\multicolumn{2}{c}{\bf{AIDATATANG}} \\
	&&\makecell[c]{\bf{dev}} &\makecell[c]{\bf{test}}&\makecell[c]{\bf{dev}} &\makecell[c]{\bf{test}} \\
	\hline 
    Hybrid &30&4.9$^\S$&5.4$^\S$&5.9$^\star$&6.7$^\star$ \\
    \hline
    Transducer&31&5.9&6.3&5.8&6.7 \\
    \hline
    CTC &26&5.7$^\S$&6.2$^\S$&5.5&6.3 \\
    Intermediate-CTC&26&5.3$^\S$&5.7$^\S$&5.1&5.9\\
	SC-CTC \cite{nozaki2021relaxing}&27&4.9&5.3&5.1&5.9\\
    \textbf{GIC (ours)}&28&\textbf{4.7}&\textbf{5.1}&\bf{4.8}&\bf{5.5}\\
  \Xhline{2\arrayrulewidth}
  \end{tabular}}
\end{table}

\begin{table}[th]
\small
\renewcommand{\arraystretch}{1.1}
  \caption{Disentangling the proposed method on AIDATA-TANG, using \textbf{Transformer} backbone. We remove the sub-layers of GIC and move towards a CTC model: (1) replacing gate units with element-wise sum operation; (2) removing the embedding layer, which is equivalent to removing the entire GIC module; (3) removing the intermediate CTC loss.}
  \label{tab:ablation}
  \centering
  \setlength\tabcolsep{3.6mm}{
  \begin{tabular}{l|cc}
  \Xhline{2\arrayrulewidth} 
	\textbf{Model Architectures}  & \textbf{Dev set} & \textbf{Test set} \\
	\hline 
    Proposed method&4.8&5.5\\
    \hspace{0.2cm}- Gate units& 5.0& 5.8 \\
    \hspace{0.4cm}- Embedding & 5.1&5.9 \\
    \hspace{0.6cm}- Intermediate CTC loss& 5.5&6.3 \\
  \Xhline{2\arrayrulewidth}
  \end{tabular}}
  \vspace{-0.2cm}
\end{table}

\subsection{Ablation Study}

The proposed approach differs from the standard CTC \cite{2006Connectionist} model in a number of ways, including the GIC block and intermediate CTC losses. We study the effect of these differences by mutating the proposed approach toward the CTC model. Table~\ref{tab:ablation} shows the impact of each change on the proposed method. 
The results demonstrate that introducing the GIC mechanism brings performance improvements from 5.1\%/5.9\% to 4.8\%/5.5\% on AIDATATANG dev/test sets.

Figure~\ref{fig:fig3} gives the CER comparison of all interlayer predictions of the SC-CTC and GIC methods. GIC achieves better performance in all intermediate layers as well as the final layer. Compared with the self-conditioning mechanism \cite{nozaki2021relaxing}, the GIC mechanism performs better to improve the performance of the CTC-based model.

\section{Conclusions}
\label{sec:conclusions}

We present an effective and novel gated interlayer collaboration (GIC) mechanism to improve the performance of the CTC-based models, which introduces the textual information into the models and thus eases the conditional independence assumption of the models. The GIC block consists of an $embedding$ layer to summarize position-specific textual representation and a gate unit to fuse the textual and acoustic features. Experiments show that our model achieves promising results both on Transformer and Conformer architectures.

\vfill\pagebreak

\clearpage
\bibliographystyle{IEEEbib}
\small
\bibliography{strings,refs}

\end{document}